\documentclass[twoside,11pt]{article}

\usepackage{jmlr2e}
\setcitestyle{numbers,square}
\usepackage{listings}
\usepackage{color}
\usepackage{CJK}
\usepackage{amsmath}

\PassOptionsToPackage{most}{tcolorbox}
\usepackage{tcolorbox}

\definecolor{dkgreen}{rgb}{0,0.6,0}
\definecolor{gray}{rgb}{0.5,0.5,0.5}
\definecolor{mauve}{rgb}{0.58,0,0.82}

\usepackage{graphicx}
\usepackage{float}
\usepackage{booktabs}
\usepackage{bm}
\usepackage{subfigure}
\usepackage[american]{babel}

\usepackage{hyperref}
\hypersetup{
    colorlinks=true,
    linkcolor=blue,
    citecolor=green,
    urlcolor=magenta
}

\newtheorem{assumption}{Assumption}

\ShortHeadings{GPT as a Monte Carlo Language Tree: A Probabilistic Perspective}{}
\firstpageno{1}

\title{GPT as a Monte Carlo Language Tree: \\A Probabilistic Perspective}
\author{\name Kun-Peng Ning \email ningkp@stu.pku.edu.cn
       \AND
       \name Jia-Yu Yao \email jiayu\_yao@pku.edu.cn 
       \AND
       \name Yu-Yang Liu \email liuyuyang13@pku.edu.cn
       \AND
       \name Mu-Nan Ning \email munanning@stu.pku.edu.cn 
       \AND
       \name Li Yuan\thanks{Coresponding Author} \email yuanli-ece@pku.edu.cn \\
      \addr 
      School of Electronic and Computer Engineering\\ 
      Peking University\\
}

\editor{}

\begin{document}

\maketitle
\vspace{-1cm}
\begin{abstract}
Large Language Models (LLMs), such as GPT, are considered to learn the latent distributions within large-scale web-crawl datasets and accomplish natural language processing (NLP) tasks by predicting the next token. 
However, this mechanism of latent distribution modeling lacks quantitative understanding and analysis. 
In this paper, we propose a novel perspective that any language dataset can be represented by a Monte Carlo Language Tree (abbreviated as ``Data-Tree''), where each node denotes a token, each edge denotes a token transition probability, and each sequence has a unique path. Any GPT-like language model can also be flattened into another Monte Carlo Language Tree (abbreviated as ``GPT-Tree''). 
Our experiments show that different GPT models trained on the same dataset exhibit significant structural similarity in GPT-Tree visualization, and larger models converge more closely to the Data-Tree. More than 87\% GPT output tokens can be recalled by Data-Tree.
These findings may confirm that the reasoning process of LLMs is more likely to be probabilistic pattern-matching rather than formal reasoning, as each model inference seems to find a context pattern with maximum probability from the Data-Tree. 
Furthermore, we provide deeper insights into issues such as hallucination, Chain-of-Thought (CoT) reasoning, and token bias in LLMs. 

\end{abstract}


\section{Introduction}
Large Language Models(LLMs), like GPT~\cite{achiam2023gpt,ouyang2022training,radford2018improving,radford2019language,brown2020language}, and LLaMA~\cite{touvron2023llama}, have demonstrated remarkable capabilities across a variety of real-world applications \cite{liu2023pre,ouyang2022training,zhao2023survey}. It is usually considered to learn the latent distributions within large-scale web-crawl datasets and accomplish natural language processing (NLP) tasks by predicting the next token \cite{zhao2024explainability}. However, this mechanism of latent distribution modeling remains unclear, and this lack of transparency poses undesirable risks for downstream applications, such as generating harmful content or hallucinations. 

\begin{figure}
    \centering
    \includegraphics[width=0.99\linewidth]{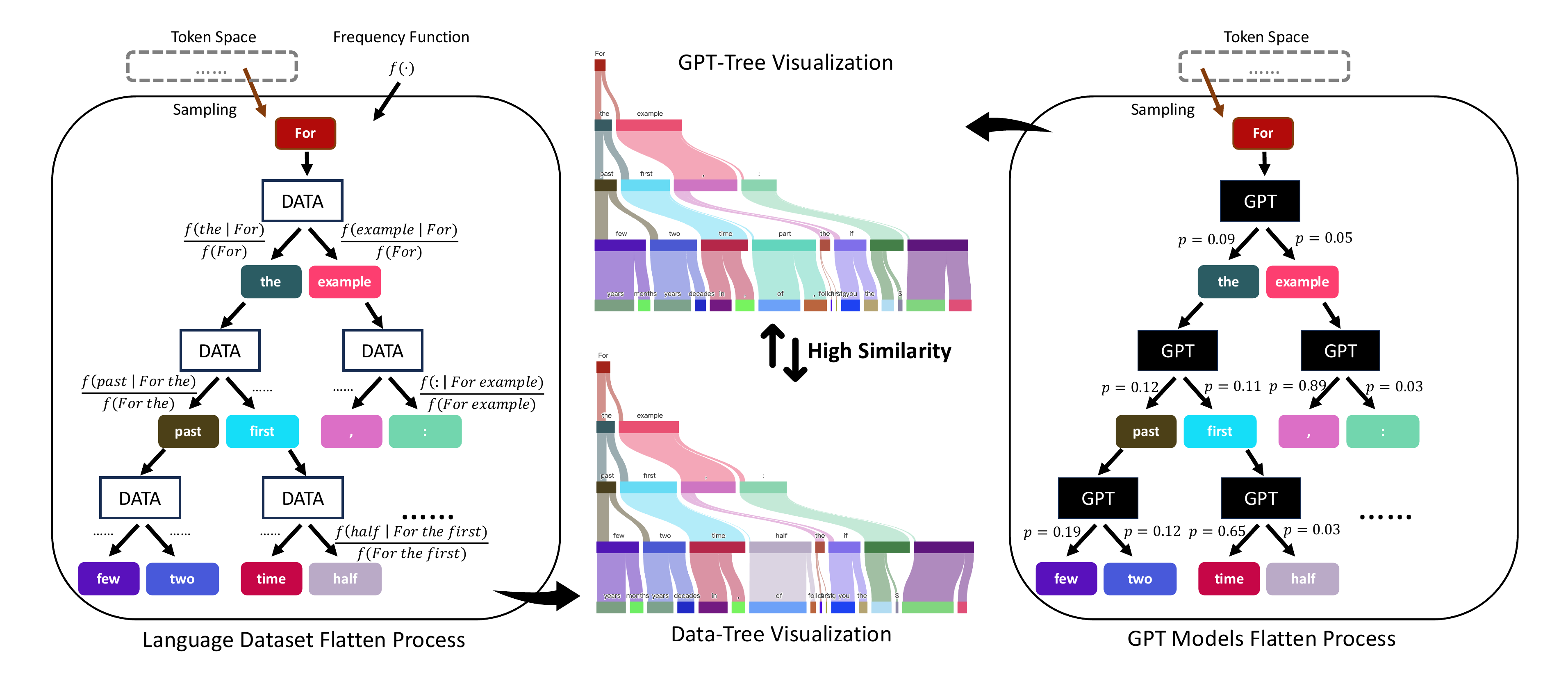}
    \caption{A novel perspective for analyzing the intelligent abilities of LLMs by representing both the language dataset and the GPT models as Monte Carlo Language Trees. }
    \label{fig.1}
\end{figure}

Most previous works \cite{deletang2023language,zhu2024survey,bellard2019lossless,chevalier2023adapting,jaiswal2023compressing} understood the intelligent abilities of LLMs from the compression perspective, \textit{i.e.}, compression for AGI. \citet{deletang2023language} shows that LLMs achieve impressive compression rates on modalities other than text, and \cite{guo2024ranking} advocates for viewing the intelligent abilities of LLMs through the lens of compression. LLM attempts to perform lossless compression on large-scale data by next-token prediction pre-training, and learns some common patterns that help downstream prediction and generalization \cite{hershcovitch2024lossless,valmeekam2023llmzip,narashiman2024alphazip}. Although reasonable, most explanations are obscure and lack quantitative understanding and analysis. 

In this paper, we propose a novel perspective for analyzing the intelligent abilities of LLMs by representing both the training data $\mathcal{D}$ and the GPT models as Monte Carlo Language Trees. As shown in Figure \ref{fig.1} (Left), we exhibit that any language dataset $\mathcal{D}$ can be perfectly represented by a Monte Carlo Language Tree (abbreviated as ``Data-Tree''), parameterized by $\theta^*$. Specifically, we sample the first token as the root node, enumerate its next token as the leaf node, and calculate the condition frequency as the edge. Repeating this process, we can obtain the ``Data-Tree'' flattened by the language dataset. On the other hand, any GPT-like language models $\hat{\theta}$ can also be flattened into another Monte Carlo Language Tree (abbreviated as ``GPT-Tree''). As shown in Figure \ref{fig.1} (Right), we randomly sample a first token (e.g., “For”) from token space, then input it into GPT and obtain the probability distribution of the next token (\textit{e.g.}, ``the'', ``example''). Next, we input “For the” and “For example” to GPT to get the next token again. Repeating this process, we can obtain the ``GPT-Tree'' flattened by GPT. 

Our results show that different language models (\textit{e.g.}, GPT-neo-X series models \cite{black2022gpt,black2021gpt}) trained on the same dataset (The Pile \cite{gao2020pile}) have very high similarity in GPT-Tree visualization (Figure \ref{fig.1} Middle, and \ref{fig:4}), and the larger the model, the closer it is to the Data-Tree $\theta^*$. More than 87\% GPT output tokens can be recalled by Data-Tree. Through our quantitative analysis, we demonstrate that using existing language models to fit training data essentially seeks a more efficient way to approximate the Data-Tree (\textit{i.e.}, $\hat{\theta} \rightarrow \theta^*$).  These findings may confirm that the reasoning process in LLMs is more likely to be \textbf{probabilistic pattern-matching} rather than formal reasoning, as each model inference seems to find a context pattern with maximum probability from the Data-Tree.

Based on this discovery, we provide deeper insights into issues such as hallucination, Chain-of-Thought (CoT) reasoning, and token bias in LLMs.
For example, our results indicate that some rare vocabulary can easily induce the model to find other incorrect GPT-Tree paths, resulting in poor response performance.

The contribution of this paper can be summarized as follows:
\begin{itemize}
    \item We propose a novel perspective for analyzing LLMs by representing the large-scale language dataset and GPT models as the Monte Calo Language Trees, named Data-Tree $\theta^*$ and GPT-Tree $\hat{\theta}$, respectively. 
    \item Quantitative analysis demonstrate that using existing language models to fit training data essentially seeks a more efficient way to approximate the Data-Tree (\textit{i.e.}, $\hat{\theta} \rightarrow \theta^*$). Furthermore, GPT-Trees generated by different language models trained on the same dataset exhibit a high degree of similarity.
    \item Our perspective of the Monte Carlo Language Tree can better explain and understand many existing counterintuitive phenomena, such as hallucinations, CoT and token bias. 
\end{itemize}

\section{Definition of Data-Tree and GPT-Tree}
In this section, we first present the definition of symbols, and then introduce the details of building Data-Tree and GPT-Tree.

\textbf{Symbols Definition.}
In this paper, we consider a large language model pre-trained scenario with a large-scale dataset $\mathcal{D}$ composed of massive sequences $x$, where $\mathcal{D}=\{x_i\}_{i=1}^{|\mathcal{D}|}$. Each sequences is consisted of variable length of tokens, denoted as $x=[t_1,t_2,...,t_n]$. Existing pre-trained methods are common to maximize the likelihood of each token $t_n$ based on its precedents $[t_1,...,t_{n-1}]$ \cite{radford2019language,brown2020language}:
\begin{equation}
\label{eq:1}
\begin{split}
    \hat{\theta}&=\mathop{\arg\max}_\theta \sum_{x \sim \mathcal{D}} p_\theta(x), \\
    s.t.\quad p_\theta(x)&=\mathop{\Pi}_{i=1}^{n}p_\theta(t_n|t_1, t_2, ...,t_{n-1}).
\end{split}
\end{equation}
Where $\hat{\theta}$ denotes the large language models pre-trained on the dataset $\mathcal{D}$.

\begin{figure}
    \centering
    \includegraphics[width=0.99\linewidth]{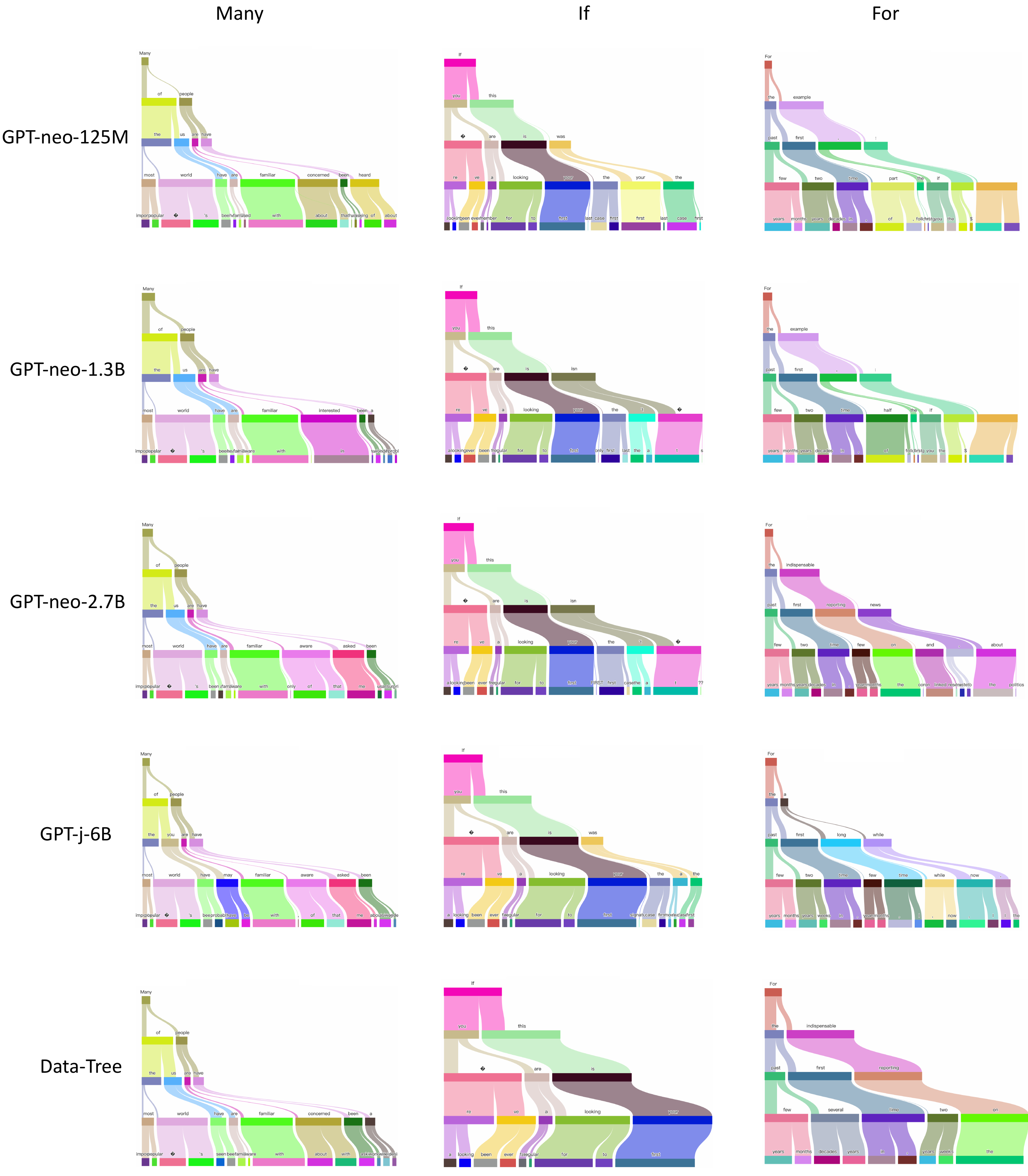}
    \caption{The tree visualization results of GPT-X series models and Data-Tree. Each row and column represents a different model (or dataset) and token, respectively. These models are trained on the same large-scale 800GB dataset, \textit{i.e.}, The Pile. Different GPT-like language models trained on the same dataset have very high similarity in GPT-Tree visualization, and gradually converge to the Data-Tree, especially on the \textbf{If} token (the second column). This similarity is not only reflected at the token level (node color of the tree), but also at the prediction probability level (edge width of the tree). }
    \label{fig:3}
\end{figure}

\textbf{Building Data-Tree.} As discussed above, any language dataset $\mathcal{D}$ can be perfectly represented by a Monte Carlo Language Tree (called ``Data-Tree''), parameterized by $\theta^*$. 
By traversing the entire dataset, we can obtain the frequency $f(t_1)$ of the first token $t_1$. We select the first token $t_1$ as the root node, enumerate its next token $t_2$ as the leaf node, and then calculate the condition frequency $p=f(t_2|t_1) / f(t_1)$ as the edge. Repeating this process, we can obtain the ``Data-Tree'' flattened by the language dataset $\mathcal{D}$. Formally, the Data-Tree $\theta^*$ satisfies the following condition:
\begin{equation}
\label{eq:2}
    p_{\theta^*}(x)=\frac{f(t_n|t_1, t_2, ...,t_{n-1})}{f(t_1, t_2, ...,t_{n-1})},\quad\forall x\sim\mathcal{D}, \forall n,
\end{equation}
where $f(\cdot)$ the frequency function, and $f(t_n|t_1, t_2, ...,t_{n-1})$ is the condition frequency of $t_n$ based on its precedents $[t_1, ..., t_{n-1}]$. \textbf{Our theoretical analysis in the Appendix \ref{theory} shows that the $\theta^*$ calculated by Eq.\ref{eq:2} is the optimal solution of Eq.\ref{eq:1}.}



\textbf{Building GPT-Tree.}
For building the ``GPT-Tree'', we also start with the first token $t_1$ sampled from token space, then input it into GPT models and obtain the probability distribution $p_{\hat\theta}(t_2|t_1)$ of the next token $t_2$. Next, we input all $[t1, t2]$ into GPT to get the next token $p_{\hat\theta}(t_3|t_1,t_2)$ again. Repeating this process, we can obtain the ``GPT-Tree'' flattened by GPT, where each node denotes a token, and each edge denotes a prediction probability $p_{\hat\theta}(\cdot)$ of the next token. Without loss of generality, for building the Data-Tree and GPT-Tree, we select one word from the letters A to Z as the first token, \textit{i.e.}, \textbf{A}s, \textbf{B}ecause, \textbf{C}ould, \textbf{D}o, \textbf{E}ven, \textbf{F}or, \textbf{G}iven, \textbf{H}owever, \textbf{I}f, \textbf{J}ust, \textbf{K}eep, \textbf{L}et, \textbf{M}any, \textbf{N}ow, \textbf{O}nce, \textbf{P}erhaps, \textbf{Q}uite, \textbf{R}ather, \textbf{S}ince, \textbf{T}he, \textbf{U}nder, \textbf{V}ery, \textbf{W}here.

\begin{figure}
    \centering
    \includegraphics[width=0.99\linewidth]{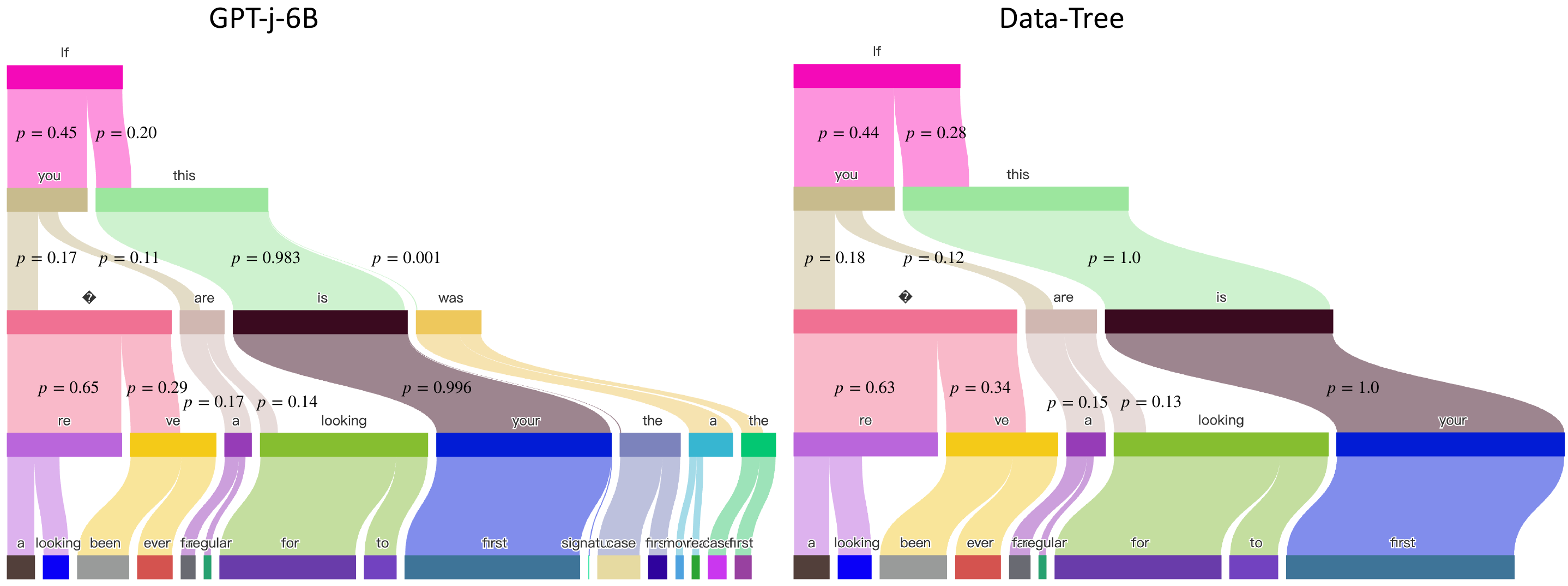}
    \caption{A closer look of visualization of GPT-Tree and Data-Tree on the ``If'' token. GPT-Tree exhibits extremely high similarity in token prediction probability with Data-Tree.}
    \label{fig:4}
\end{figure}

\section{Understanding with Perspective of Monte Carlo Language Tree}

\subsection{The GPT Models Gradually Converge to the Data-Tree}

To evaluate our main idea, we conduct experiments on \textbf{The Pile}\footnote{https://pile.eleuther.ai/} dataset \cite{gao2020pile}, which is one of the largest open-source large-scale language datasets, with approximately 825GB. We employ the following four LLMs for further analysis, \textit{i.e.}, GPT-neo-125M, GPT-neo-1.3B, GPT-neo-2.7B \cite{black2021gpt}, and GPT-j-6B \cite{wang2021gpt}, all of which are trained on The Pile dataset \footnote{https://huggingface.co/datasets/EleutherAI/pile}. 

We first plot the tree visualization of GPT-neo-X series and GPT-j-6B models in Figure \ref{fig:3}, all of which are trained on the same large-scale 800GB dataset, \textit{i.e.}, The Pile. Additionally, the Data-Tree visualization of The Pile is also appended in Figure \ref{fig:3}. Each row and column represents a different model (or dataset) and token respectively, and more results can be found in the Appendix.  

As shown in Figure \ref{fig:3}, it can be observed that different GPT-like language models trained on the same dataset have very high similarity in GPT-Tree visualization, and gradually converge to the Data-Tree, especially on the \textbf{If} token (the second column). This similarity is not only reflected at the token level (node color of the tree), but also at the prediction probability level (edge width of the tree). It is worth noting that the \textbf{left half of the tree} has higher similarity, as these are the parts with higher prediction probabilities. GPT-Trees gradually become dissimilar as depth increases and prediction probability decreases, due to the increasing uncertainty they generate. 

From a closer look at Figure \ref{fig:4}, we draw the probability values of each edge for the GPT-Tree and Data-Tree. We can find the learned $p_{\hat{\theta}}(\cdot)$ is gradually approaching $p_{\theta^*}(\cdot)$, for example, $\{p_{\hat{\theta}}(you|If)=0.45,p_{\theta^*}(you|If)=0.44\}$, $\{p_{\hat{\theta}}('|If,you)=0.17,p_{\theta^*}('|If,you)=0.18\}$, $\{p_{\hat{\theta}}(are|If,you)=0.11,p_{\theta^*}(are|If,you)=0.12\}$, $\{p_{\hat{\theta}}(is|If,this)=0.983,p_{\theta^*}(is|If,this)=1.0\}$, $\{p_{\hat{\theta}}(your|If,this,is)=0.996,p_{\theta^*}(your|If,this,is)=1.0\}$, and so on. These findings once again verify our idea that the GPT models gradually converge to the Data-Tree.

\begin{figure}
    \centering
    \includegraphics[width=0.48\linewidth]{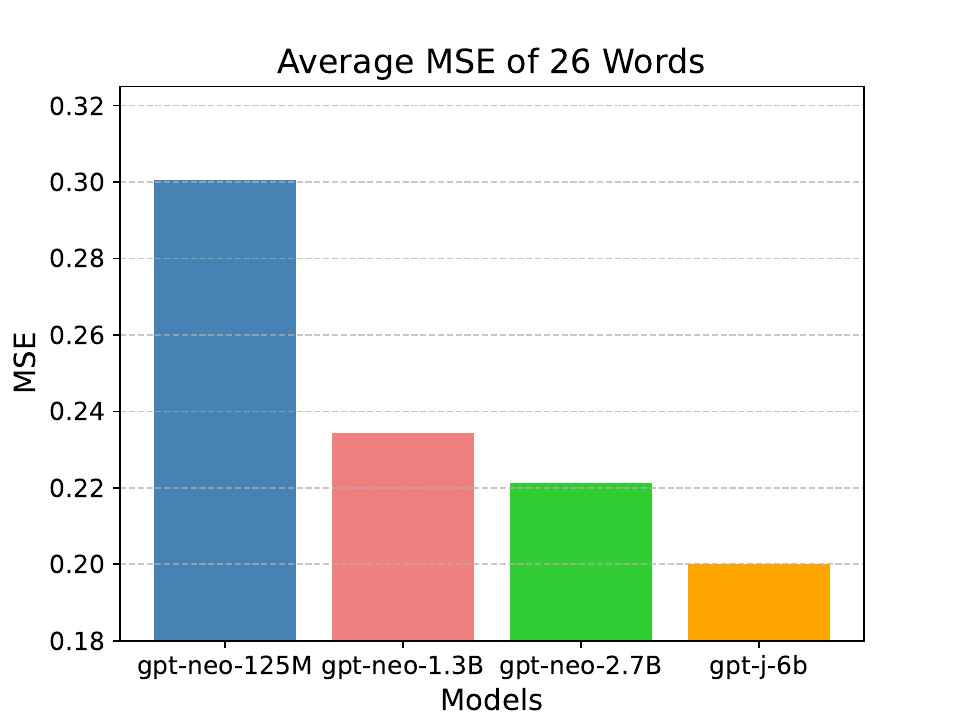}
    \includegraphics[width=0.48\linewidth]{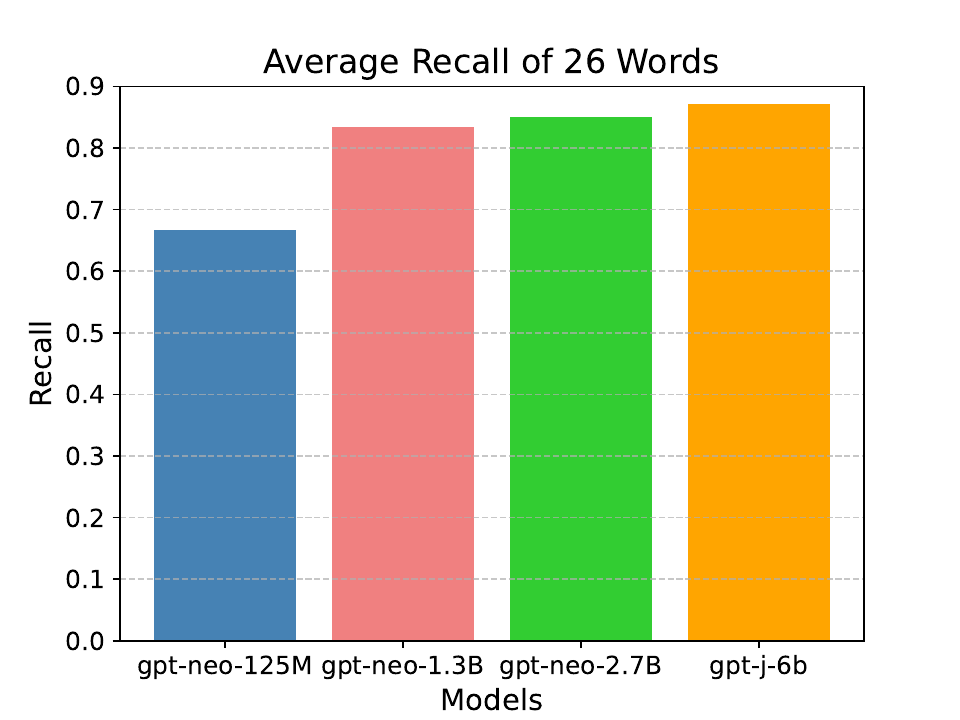}
    \caption{Average MSE and Recall@5 of 26 words. The MSE measures the probability difference between the GPT-Tree and Data-Tree, the smaller the value, the more similar it is. The Recall@5 measures how many GPT output tokens will be recalled by Data-Tree. The results show that the GPT models gradually converge to the Data-Tree with increased parameter size.}
    \label{fig:5}
\end{figure}

To further quantify this finding, we show the statistical results of MSE and Recall@5 (as shown in Eq.\ref{eq:3} and \ref{eq:4}) in Figure \ref{fig:5}. It can be observed that the MSE results decrease as the model size increases, the larger the model, the closer it is to the Data-Tree. On the other hand, from the recall@5 results in Figure 5, more than 87\% GPT output tokens will be recalled by Data-Tree. And this recall also increases with the increase of the model parameters. 

\textbf{Result Analysis.}  Our findings reveal that employing existing language models $\hat{\theta}$ to fit the training data is essentially finding a more efficient way to approximate the Data-Tree $\theta^*$, \textit{i.e.}, $\hat{\theta}\rightarrow\theta^*$. As the parameters of the model increase, the fitting ability becomes stronger, and the GPT-Tree becomes more and more like Data-Tree. These results further confirm that the reasoning process in LLMs is more likely to be \textbf{probabilistic pattern-matching} rather than formal reasoning, as each model inference is to find a maximum probability path on this GPT-Tree. Based on this perspective (\textit{i.e.}, GPT is similar to Data-Tree), we found that it can better explain and understand many existing counterintuitive phenomena, like token bias, hallucination, and other issues. Therefore, we will discuss these in detail below.

\subsection{Understand the Token Bias and Hallucination Phenomena from GPT-Tree}
The \textit{token-bias} issues in LLMs are first proposed in \cite{jiang2024peek} and further studied in \cite{mirzadeh2024gsm}. Small changes in input tokens can drastically alter model outputs, indicating a strong token bias and suggesting that these models are highly fragile and do not possess genuine reasoning abilities \cite{jiang2024peek,shi2023large,kambhampati2024can,schaeffer2024emergent}. On the other hand, the hallucination problem also troubles us, as the models fabricate non-existent facts to deceive users with or without their awareness \cite{yao2023llm}. In this subsection, we attempt to understand why these phenomena occur from the perspective of GPT-Tree. 

\begin{figure}
    \centering
    \includegraphics[width=0.48\linewidth]{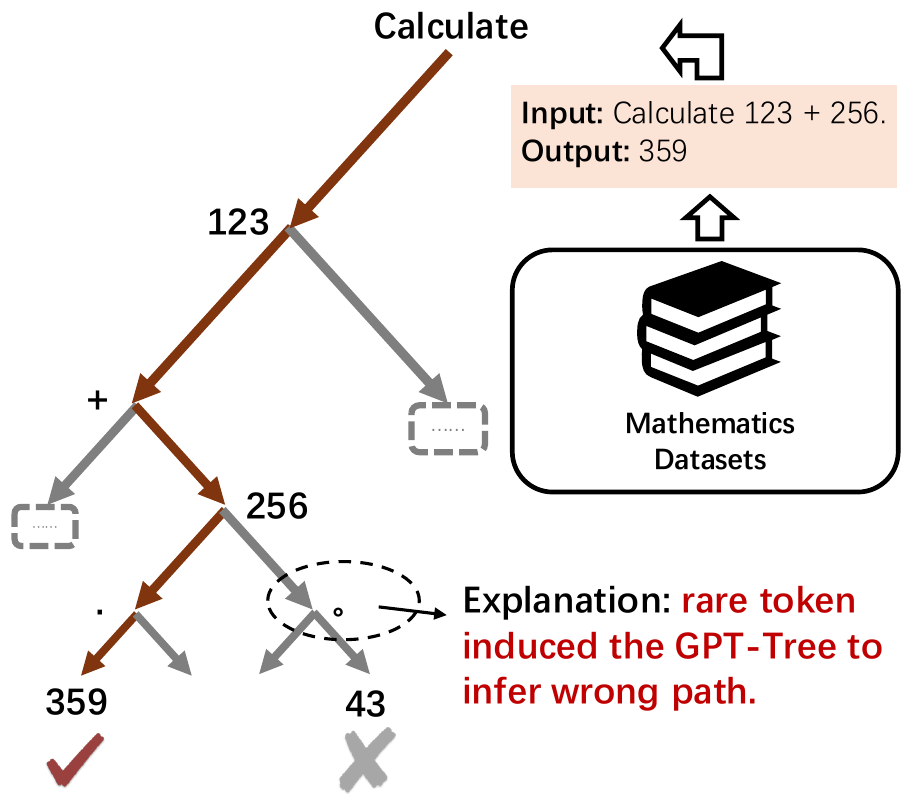}
    \includegraphics[width=0.48\linewidth]{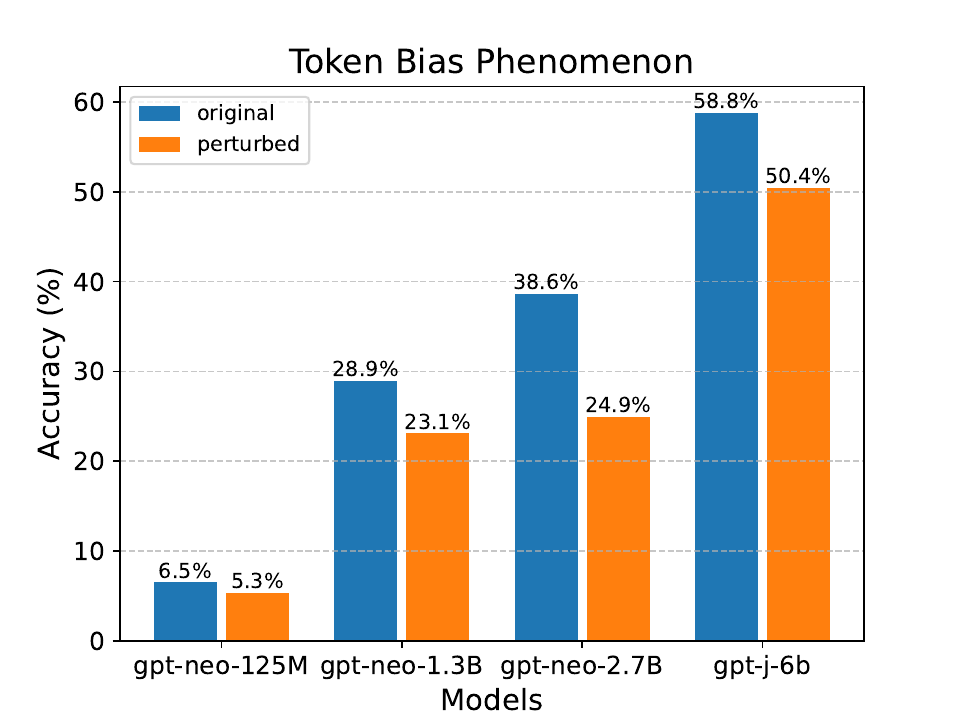}
    \caption{Understanding token bias issues in LLMs. \textbf{(Left):} The mathematical dataset contains a large number of patterns, like ``Calculate \{\%f\} \{op\} \{\%f\}.\textbackslash n\{\%f\}''. We sample the original question (\textit{``Calculate 123 + 256.''}) from the mathematical dataset, the GPT model could answer \textit{``359''} correctly. However, we perturb the last token ``.'' into ``\begin{CJK}{UTF8}{gbsn}。\end{CJK}'', the model incorrectly answers as \textit{``43''}.  We suggest that token bias is due to some rare tokens inducing the GPT-Tree to infer the wrong path. \textbf{(Right):} We further quantified this phenomenon by evaluating the original (blue bars) and perturbed (orange bars) accuracy of different models in 21076 QA test pairs. The accuracy of all models has significantly decreased after perturbing the last token ``.'' into ``\begin{CJK}{UTF8}{gbsn}。\end{CJK}'', suffering from the token bias issues.}
    \label{fig:6}
\end{figure}

\begin{figure}
    \centering
    \includegraphics[width=0.95\linewidth]{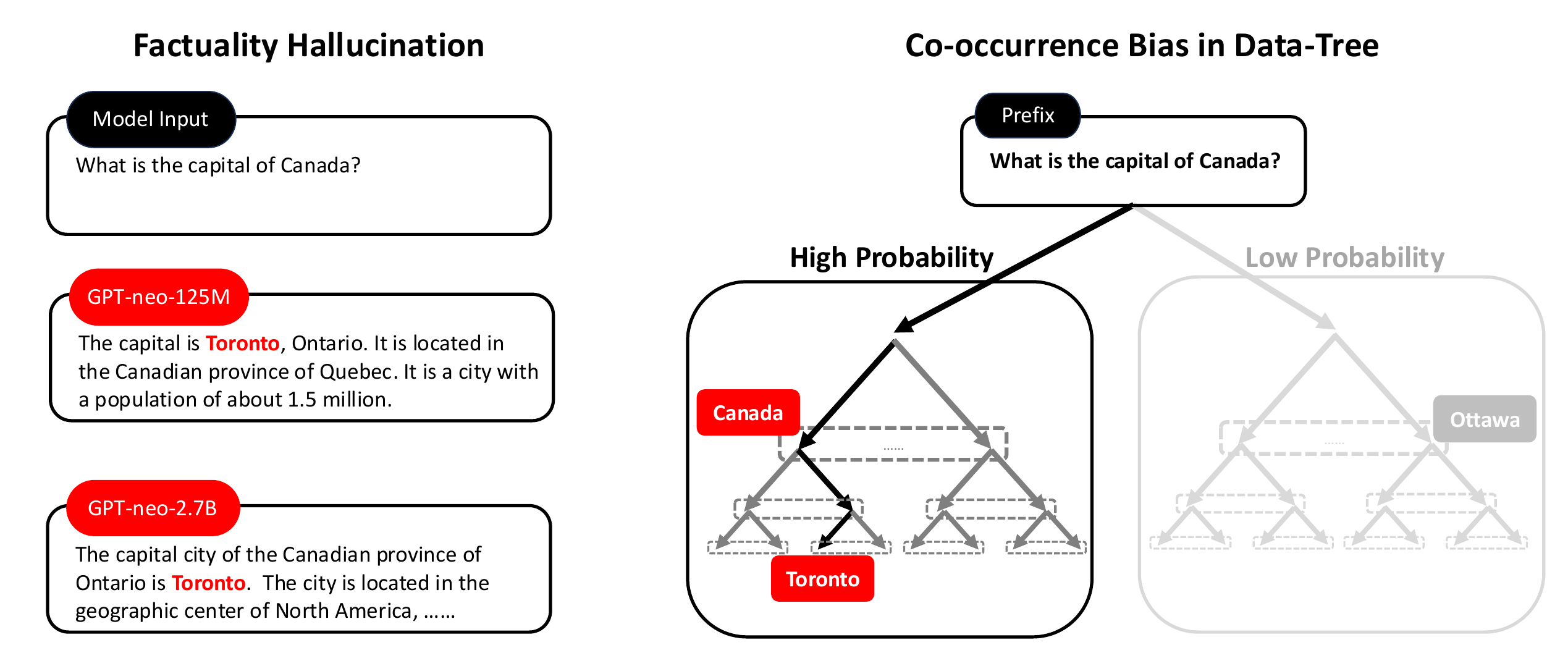}
    \caption{Understanding Hallucination Phenomenon. \textbf{(Left):} Input ``What is the capital of Canada?'' into the model, and both GPT-neo-125M and GPT-neo-2.7B models answer ``The capital is \textbf{\textcolor{red}{Toronto}}'', exhibiting factuality hallucination. \textbf{(Right):} The GPT model leans heavily on the frequent co-occurrence of the terms \textit{Toronto} and \textit{Canada} in its training data, without truly capturing the factual knowledge.}
    
    \label{fig:7}
\end{figure}

\textbf{Understanding Token Bias Issues.} We note that the Pile contains a mathematics dataset \cite{saxton2019analysing} that includes 2 million mathematical QA pairs. The example questions are as follows. 
\begin{tcolorbox}[colback=gray!20!white, colbacktitle=white, coltitle=black, colframe=black!75!black, boxrule=0.7pt, halign title=center]
\textbf{Question:} Calculate -5*39/(-130) - (-20)/(-8).

\textbf{Answer:} -1
\\\\
\textbf{Question:} Calculate 42/60*40/70.

\textbf{Answer:} 2/5
\\\\
\textbf{Question:} Calculate -841880142.544 + 411127.

\textbf{Answer:} -841469015.544
\end{tcolorbox}
Among them, there are 21076 QA formats as follows, ``Calculate \{\%f\} \{op\} \{\%f\}.\textbackslash n\{\%f\}''. Where \{\%f\} denotes different float values, and \{op\} refers to one of the $\{+,-,*,/\}$. As show in Figure \ref{fig:6} (Left), we sample the original question (\textit{``Calculate 123 + 256.''}) from the mathematical dataset, and the GPT model could answer \textit{``359''} correctly. However, we perturb the last token ``.'' into ``\begin{CJK}{UTF8}{gbsn}。\end{CJK}'', the model incorrectly answers as \textit{``43''}.  We suggest that \textbf{token bias is due to some rare tokens inducing the GPT-Tree to infer the wrong path}. Additionally, we further quantified this phenomenon in Figure \ref{fig:6} (Right). We evaluated the original (blue bars) and perturbed (orange bars) accuracy of different models in the above 21076 QA pairs. It can be observed that the accuracy of all models has significantly decreased after perturbing the last token ``.'' into ``\begin{CJK}{UTF8}{gbsn}。\end{CJK}'', especially on GPT-neo-2.7B ($38.6\%\rightarrow24.9\%$).

\textbf{Understanding Hallucination Phenomenon.} LLMs suffer from hallucinations, fabricating non-existent facts to deceive users. As shown in Figure \ref{fig:7} (Left), we input ``What is the capital of Canada?'' into the model, and both GPT-neo-125M and GPT-neo-2.7B models answer ``The capital is \textbf{\textcolor{red}{Toronto}}.'' We believe it is caused by the co-occurrence bias of the Data-Tree. As shown in Figure \ref{fig:7} (Right), the training data exhibit high-frequency co-occurrence of the terms \textit{Toronto} and \textit{Canada}, while the GPT model leans heavily towards these corpora, without truly capturing the factual knowledge about the capital of Canada. Similar conclusions are discussed in \cite{huang2023survey,yao2023llm}.

\begin{figure}
    \centering
    \includegraphics[width=0.95\linewidth]{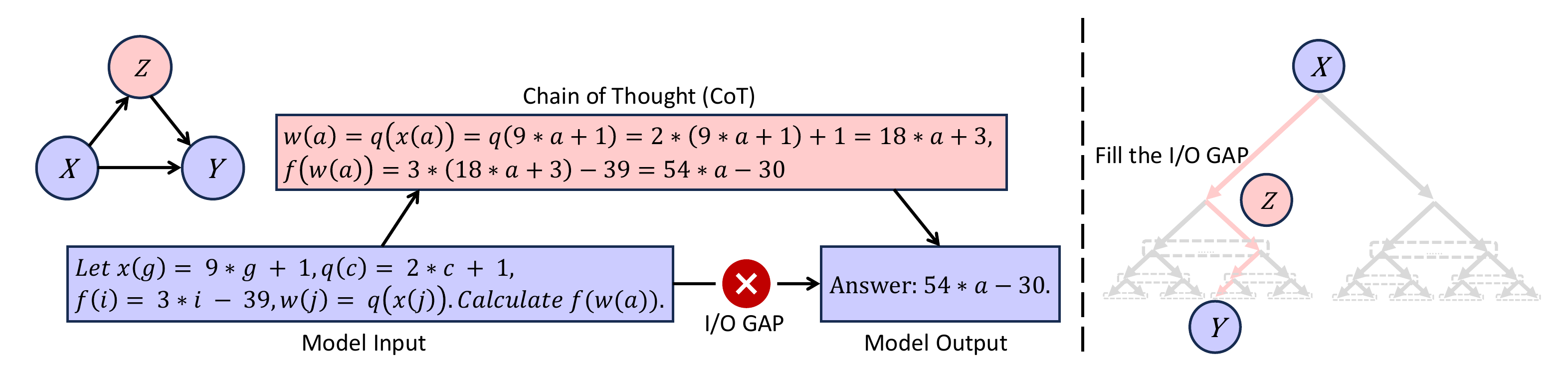}
    \caption{Understanding the effectiveness of Chain-of-Thought (CoT) from GPT-Tree. \textbf{(Left):} For some complex issues, there is a significant gap between input $X$ and output $Y$, making it difficult for the GPT model to directly derive $Y$ from $X$. \textbf{(Right):} From the perspective of GPT-Tree, the input $X$ is at a parent node, and the output $Y$ is at a deep leaf node. The CoT focuses on finding a path $Z$ to help the GPT model connect $X$ and $Y$, attempting to fill the I/O gap.}
    \label{fig:8}
\end{figure}

\subsection{Understand the Effectiveness of Chain-of-Thought (CoT) from GPT-Tree}
From the perspective of GPT-Tree, we can better explain and understand the effectiveness of the well-known Chain-of-Thought (CoT) \cite{wei2022chain}. As shown in Figure \ref{fig:8} (Left), for some complex issues, there is a significant I/O gap between input $X$ and output $Y$, making it difficult for the GPT model to directly derive $Y$ from $X$. Numerous studies \cite{wei2022chain,yao2024tree,long2023large,besta2024graph,zheng2023judging,naveed2023comprehensive,liang2023encouraging,wang2022self} have shown that introducing the thought processes (\textit{e.g.}, CoT \cite{wei2022chain}, ToT \cite{yao2024tree}, GoT \cite{besta2024graph}) can enable the GPT model to output more accurate responses, especially on mathematical tasks \cite{frieder2024mathematical}. However, few studies explain why it works. In our GPT-Tree perspective, we believe that the CoT induces the GPT model to output the thought process $Z$, which can fill the I/O gap between the input $X$ and output $Y$. As shown in Figure \ref{fig:8} (Right), the input $X$ is at a parent node of the GPT-Tree, and the output $Y$ is at a deep leaf node. \textbf{The CoT attempts to find a path $Z$ to help the GPT model connect $X$ and $Y$}.


\section{Conclusion}
This paper studies the properties of large language models from a novel perspective. Specifically, we flatten the large-scale language dataset and GPT-like model into the Monte Calo Language Trees, named Data-Tree and GPT-Tree, respectively. Each node denotes a token, each edge denotes a token transition probability, and each sequence has a unique path. Our quantitative and qualitative results show that different GPT models trained on the same dataset have high similarity in GPT-Tree visualization, and the larger the model, the closer it is to the Data-Tree. In other words, employing existing language models to fit the training data is essentially finding a more efficient way to approximate the Data-Tree, \textit{i.e.}, the GPT model is gradually converging to the Data-Tree. These findings further confirm that the reasoning process in LLMs is more likely to be probabilistic pattern-matching rather than formal reasoning, as each model inference is to find a maximum probability path on this GPT-Tree. Additionally, our perspective (\textit{i.e.}, GPT is similar to Data-Tree) can better explain and understand many existing counterintuitive phenomena, like token bias, hallucination, and chain-of-thought. 

\vskip 0.2in
\clearpage
\bibliography{gpt}

\clearpage

\appendix




\section{Theory Analysis}
\label{theory}

\begin{assumption}[Ideal Conditions]\label{assum:ideal}
  The following conditions hold:
  \begin{enumerate}
    \item The model has unbounded capacity (no parametric restrictions);
    \item The training dataset size approaches infinity, with each context $h$ appearing infinitely often ($N(h) \to \infty$);
    \item The optimization converges to the global optimum.
  \end{enumerate}
\end{assumption}

\begin{theorem}[GPT Models Gradually Converge to the Data-Tree]\label{thm:main}
  Under Assumption \ref{assum:ideal}, the optimal conditional probability distribution $p_{\theta^*}(w|h)$ learned by an autoregressive language model through maximum likelihood estimation (Eq.\ref{eq:1}) equals the conditional frequency $\hat{p}(w|h)$ (Eq.\ref{eq:2}) in the training dataset:
  \begin{equation}
    p_{\theta^*}(w|h) = \hat{p}(w|h) \triangleq \frac{N(h,w)}{N(h)}
  \end{equation}
  where $N(h)$ counts occurrences of context $h$, and $N(h,w)$ counts co-occurrences of context $h$ with the next token $w$.
\end{theorem}

\begin{proof}
  Let the training dataset $\mathcal{D} = \{(h_i, w_i)\}_{i=1}^N$ consist of i.i.d. samples, where $h_i$ denotes the context and $w_i$ the next token. Thus we can define:
  \begin{align}
    N(h) &\triangleq \sum_{(h',w') \in \mathcal{D}} \mathbb{I}(h' = h) \\
    N(h,w) &\triangleq \sum_{(h',w') \in \mathcal{D}} \mathbb{I}(h' = h \land w' = w)
  \end{align}
  
  The model parameters $\theta$ are learned by maximizing the log-likelihood:
  \begin{equation}
    \mathcal{L}(\theta) = \sum_{(h,w) \in \mathcal{D}} \log p_\theta(w|h)
  \end{equation}
  
  For each fixed context $h$, this decomposes into independent optimization problems:
  \begin{equation}
    \max_{p(\cdot|h)} \sum_w N(h,w) \log p(w|h) \quad \text{s.t.} \quad \sum_w p(w|h) = 1
  \end{equation}
  
  Form the Lagrangian:
  \begin{equation}
    \mathcal{J} = \sum_w N(h,w) \log p(w|h) + \lambda\left(1 - \sum_w p(w|h)\right)
  \end{equation}
  
  Taking derivatives with respect to $p(w|h)$:
  \begin{equation}
    \frac{\partial \mathcal{J}}{\partial p(w|h)} = \frac{N(h,w)}{p(w|h)} - \lambda = 0 \quad \Rightarrow \quad p(w|h) = \frac{N(h,w)}{\lambda}
  \end{equation}
  
  Enforcing the constraint $\sum_w p(w|h) = 1$:
  \begin{equation}
    \sum_w \frac{N(h,w)}{\lambda} = 1 \quad \Rightarrow \quad \lambda = \sum_w N(h,w) = N(h)
  \end{equation}
  
  Thus the optimal solution is:
  \begin{equation}
    p_{\theta^*}(w|h) = \frac{N(h,w)}{N(h)} = \hat{p}(w|h) \quad \blacksquare
  \end{equation}
\end{proof}


\section{Experimental Details}

\subsection{Experiments and Results}

In this section, we present our main results and postpone complementary findings to the Appendix. To evaluate our main idea, we conduct experiments on \textbf{The Pile}\footnote{https://pile.eleuther.ai/} dataset \cite{gao2020pile}, which is one of the largest open-source large-scale language datasets, with approximately 825GB. We employ the following four LLMs for further analysis, \textit{i.e.}, GPT-neo-125M, GPT-neo-1.3B, GPT-neo-2.7B \cite{black2021gpt}, and GPT-j-6B \cite{wang2021gpt}, all of which are trained on The Pile dataset \footnote{https://huggingface.co/datasets/EleutherAI/pile}. 

\textbf{Evaluation Metrics.} This paper verifies our idea in three ways. \textbf{\textit{i) Tree Visualization:}} For each first token, we use the echart's Sankey diagram \footnote{https://echarts.apache.org/examples/zh/index.html\#chart-type-sankey} to visualize the Data-Tree and GPT-Tree. The same color represents the same token, and the width of the edge represents the probability of the next token. For each layer of the tree, the probability of the left leaf node is greater than the right one. \textbf{\textit{ii) Mean Squared Error:}} We use the mean squared error (MSE) to measure the alignment between the GPT-Tree $\hat{\theta}$ and Data-Tree $\theta^*$, by calculating the probability difference of each edge, 
\begin{equation}
\label{eq:3}
    MSE=\frac{1}{N_T^K}\sum_{[t_1,...,t_n]}[p_{\theta^*}(t_n|t_1,...t_{n-1})-p_{\hat{\theta}}(t_n|t_1,...t_{n-1})]^2,
\end{equation}
where $N_T^K$ denotes the total number of tokens on the tree with a depth of $T$ for Top-$K$ flattening. \textbf{\textit{iii) Recall@5:}} We use this metric to measure how many GPT $\hat{\theta}$ output tokens will be recalled by Data-Tree $\theta^*$,
\begin{equation}
\label{eq:4}
    Recall@5=\frac{\sum_{[t_1,...,t_n]}\mathbf{1}\{\hat{t}_n \in Top5[p_{\theta^*}(\cdot|t_1,...t_{n-1})]\}}{N_T^K}
\end{equation}
where $\hat{t}_n=\max p_{\hat\theta}(\cdot|t_1,...t_{n-1})$ denotes the maximum predicted token by GPT-Tree, and $Top5[p_{\theta^*}(\cdot|t_1,...t_{n-1})]$ denotes the Top-$5$ next tokens from Data-Tree.

\subsection{Detail of Building Data-Tree} 
The key of building the Data-Tree is how to partition appropriate data chunks to calculate Eq.\ref{eq:2}. From the GPT-3 technical report \cite{brown2020language}, we can find the details of data partitioning in Appendix B and C, \textit{i.e.}, ``\textit{During training we always train on sequences of the full 2048 token context window, packing multiple documents into a single sequence when documents are shorter than 2048, pieces less than 200 characters long were discarded.}''. Additionally, we notice that both the GPT-neo-X \cite{black2021gpt,black2022gpt} series and GPT-j-6B \cite{wang2021gpt} models replicate the experimental setup of GPT-3. Therefore, we also follow this data partitioning method, \textit{i.e.}, splitting each document into 2048 size slices and discard data chunks smaller than 200. For each data chunk, we use the BPE-based tokenizer used in the GPT-neo-X series, GPT-j-6B, and GPT3 models. After that, we select the first token $t_1$ as the root node of the Data-Tree, enumerate its next token $t_2$ as the leaf node, and then calculate the condition frequency $p=f(t_2|t_1) / f(t_1)$  from all data chunks as the edge. Repeating this process, we can obtain the ``Data-Tree'' started with $t_1$ flattened by the Pile dataset.

\end{document}